# Spiking Neural Networks for Early Prediction in Human Robot Collaboration

Tian Zhou and Juan P. Wachs

*Abstract* —This paper introduces the Turn-Taking Spiking Neural Network (TTSNet), which is a cognitive model to perform early turn-taking prediction about human or agent's intentions. The TTSNet framework relies on implicit and explicit multimodal communication cues (physical, neurological and physiological) to be able to predict when the turn-taking event will occur in a robust and unambiguous fashion. To test the theories proposed, the TTSNet framework was implemented on an assistant robotic nurse, which predicts surgeon's turn-taking intentions and delivers surgical instruments accordingly. Experiments were conducted to evaluate TTSNet's performance in early turn-taking prediction. It was found to reach a $F_1$ score of 0.683 given 10% of completed action, and a $F_1$ score of 0.852 at 50% and 0.894 at 100% of the completed action. This performance outperformed multiple state-of-the-art algorithms, and surpassed human performance when limited partial observation is given (< 40%). Such early turn-taking prediction capability would allow robots to perform collaborative actions proactively, in order to facilitate collaboration and increase team efficiency.

## I. INTRODUCTION

Turn-taking is a key component in interpersonal collaboration. It determines the timing, the roles and the basic structure in scenarios such as conversations (Sacks et al., 1974), group problem-solving (Inkpen et al., 1997) and shared control (Chan et al., 2008). In the course of a collaboration, each participating agent needs to analyze the task in progress and the ongoing communication cues, in order to determine whether, when and how to take the incoming turn. In its most fundamental configuration, the turn-taking process is defined by two agents and a task, where each agent takes turns to work on the collaborative task. Uncoordinated turn-takings will result in transitions with long gaps, overlaps and conflicts, breaking the collaboration flow (see Figure 1). A fluent, natural and coupled turn-taking process can enhance collaboration efficiency, (Sebanz et al., 2006), improve task performance (Inkpen et al., 1997; Oren et al., 2012) and strength communication grounding among team members (Marsh et al., 2009). In high-risk and high-paced tasks, such as surgery, effective turn-taking is key to the task success. Even when observing simple tasks, such as the exchanging surgical instruments, one can appreciate smooth, fluent and precise turn-taking coordination. For this reason, work in the Operating Room (OR) was chosen as the test-bed for the framework presented.

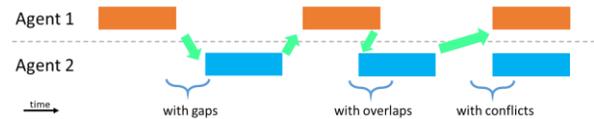

Figure 1. Illustration of a turn-regulation process between two agents

The same turn-taking norms in human-human interaction are also expected in human-robot interaction scenarios. When collaborating with humans, robots are expected to understand human partner's turn-taking intentions and the right timing to engage in an interaction. In the context of OR, Robotic Scrub Nurses (RSN) are being built to manage, deliver and retrieve surgical instruments to/from surgeons (Zhou and Wachs, 2017, 2016), as shown in Figure 2. The RSN system is anticipated to understand surgeon's implicit communication cues (e.g., change of body posture), explicit communication cues (e.g., uttering the word "scalpel") and current task progress (Chao and Thomaz, 2012). However, collaborative robots in general, lack the competence to reason about human's turn-taking intentions in a correct, robust and proactive manner. This paper fills this gap by proposing a framework in which robots can reason about human turn-taking intentions. Different dimensions of the turn-taking action are covered within this framework, including decisions about whether or not humans want to relinquish the turn, the early timing of the incoming turn-switch action, and what objects are being expected by humans in the next turn.

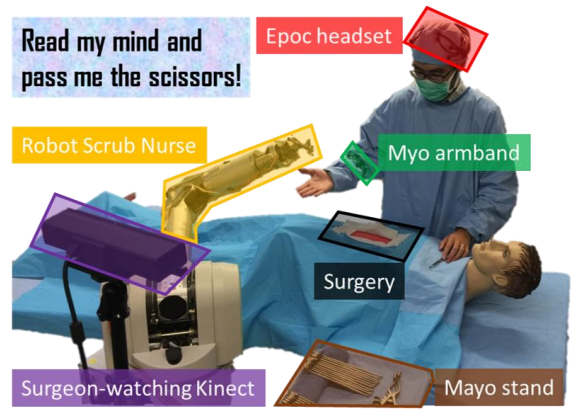

Figure 2. Illustration of a RSN system. Various sensors (Kinect, Epoc and Myo) are used to capture human's communication cues, which are fed into the TTSNet framework for turn-taking analysis. The netwok output are precise inferences about the delivery of surgical instruments ahead of time.

* Research supported by NPRP award (NPRP 6-449-2-181) from the Qatar National Research Fund (a member of The Qatar Foundation). The statements made herein are solely the responsibility of the authors.

Tian Zhou is a PhD candidate with the School of Industrial Engineering at Purdue University, West Lafayette, IN 47906 USA (e-mail: zhou338@purdue.edu).

Juan P. Wachs is an Associate Professor with the School of Industrial Engineering at Purdue University, West Lafayette, IN 47906 USA (corresponding author, phone: 765 496-7380; fax: 765 494-1299; e-mail: jpwachs@purdue.edu).

Surgeon-nurse teaming is a type of asymmetric collaboration, where the surgeon leads the task (i.e., a dominant agent) while the nurse mainly follows the task (i.e., a submissive agent). In this scenario, the focus is on enabling the *follower* to predict the *leader*'s turn-taking intention to collaborate efficiently. Thus, this paper focuses on developing frameworks to enable RSN to predict surgeons' turn-taking intentions. A typical process when human relinquishes his turn to the robot is shown in Figure 3. As human is getting towards the end of his turn, he starts revealing implicit cues (i.e., physiological and physical cues) and then explicit cues (i.e., utterances) to reflect that intention. These multimodal communication signals together indicate the human's willingness to give out the turn. The robot, in the meantime, captures those subtle cues and spots the end of human's turn. The earlier the robot can recognize human's turn-giving intention, the earlier it can start performing preparatory actions to ease the turn transition procedure.

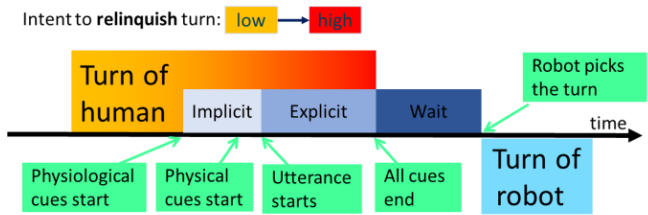

Figure 3. Illustration of the process when human transits the turn to the robot. The color change from yellow to red indicates an increasing level of human intent to relinquish his/her turn.

Research has been conducted for human's turn-taking intention recognition. In the area of human computer interaction, conversational turn-taking has been studied to help virtual agents to determine the right timing to engage in conversation (DeVault et al., 2015). In human robot interaction, physical turn-taking has been investigated in manufacturing floor (Tan et al., 2009) and robotic companions (Chao and Thomaz, 2016). However, current turn-taking recognition algorithm build on mathematically derived machine learning models and lack cognitive reasoning capabilities. For example, Support Vector Machines (SVM) (Arsikere et al., 2015), Decision Trees (DT) (Saito et al., 2015) and Conditional Random Field (CRF) (De Kok and Heylen, 2009) have been used to recognize the end-of-turn in human's conversations. Even though these algorithms can reach a certain level of recognition accuracy, they are still far from human-like competence level (Heeman and Lunsford, 2015). Furthermore, these turn-taking models are derived computationally and mathematically, and the resultant behaviors cannot be explained and interpreted well by humans. The fundamental model structure and the underlying reasoning process are different from those shown by humans. Hence, a cognitive-based turn-taking reasoning model is required. This model should reach a similar human-level competence in recognition accuracies, and should be easily interpreted by humans.

This paper introduces the Turn-Taking Spiking Neural Networks (TTSNet), which has the capability of predicting human's turn-taking intentions early on, with high accuracy and robustness. The TTSNet has biologically-inspired Spiking Neural Network (SNN) (Maass, 1997) as its core to model turn-taking processes, and several machine-learning algorithms as peripherals to help interface with the multimodal input/output signals. The TTSNet framework can distinguish signature turn-taking patterns from multimodal human behavior, and reason about human's intentions to keep or relinquish the turn in the near future. One advantage of TTSNet is that, by incorporating SNN as its basis component, it can deal with asynchronous signals in multimodal turn-taking.

Compared to traditional Artificial Neural Network (ANN) which dictates a fixed time for signal passing between neurons, SNN adds timing simulation by using inter-neuron edges of different lengths. These different lengths lead to variable traversing times (i.e., due to the variable axonal conduction delays) (Maass, 1997). When a neuron fires in SNN, it produces a signal that propagates to the connecting neurons leading to a series of neurons to fire together (i.e., causing a spike train). Such spike trains together form a polychronous neuronal groups (PNGs), which refers to a group of neurons fired together in a time-locked pattern, after being triggered by a specific input pattern. PNGs can form a rich representation of input spatio-temporal signals, and can be used as a salient feature for pattern classification. Due to this temporal modeling capability, SNN has been shown to be effective in modeling time-sensitive sequences, such as gesture recognition (Botzheim et al., 2012), speech recognition (Loiselle et al., 2005) and seizure detection (Ghosh-Dastidar and Adeli, 2007).

The proposed TTSNet framework is evaluated in the context of OR, where an RSN system needs to predict surgeon's turn-taking intentions and then perform actions accordingly. To enable multimodal sensing, different sensors were applied to capture surgeons' behavior (represented as signals). Those signals were fed to the TTSNet framework to guide the movement of the robot. The TTSNet performance is evaluated on a multimodal human behavior corpus that was collected in the laboratory environment, as a prerequisite for moving into real ORs for clinical validation.

The rest of the paper is organized as the following. Section II, we provide an overview of related work. Section III is dedicated to defining the turn-taking problem. Section IV introduces the background knowledge of SNN and section V explains the TTSNet framework in detail. The experiments are presented in section VI, followed by discussions of the observed results in section VII. Last, section VIII summarizes the paper with concluding remarks and future work.

## II. RELATED WORK

In this section, we give an overview of the related work about turn-taking analysis, with a focus on psychology research about turn-taking (section II.A), conversational turn-taking (section II.B), turn-taking in embodied agents (section II.C) and collaborative turn-taking in physical tasks (section II.D).

### A. Turn-taking in psychology

Turn-taking, as a fundamental human behavior, has been studied by cognitive scientists in the last forty years. Turn-taking routines are found to be an essential part in mother-infant gaze interactions, and deviations from the expected turn-taking process was found to lead to increased anxiety in infants (Trevarthen, 1979). In the context of problem solving among children, different turn-taking strategies were compared and it was found that the level of achievement was highly dependent on turn-taking strategies adopted (Inkpen et al., 1997).

A comprehensive overview of turn-taking studies from the psychological perspective can be found in (Holler et al., 2016). All these works reinforce the concept that turn-taking is a natural and fundamental behavior among humans, and has a great impact on emotions and objective task performance.

### B. Conversational turn-taking

The vast majority of turn-taking research came from the linguistics field, especially in the field of conversational analysis. Linguistic structures, semantics and syntax are necessary to understand turns in conversations. A seminal work by Sacks et al. (Sacks et al., 1974) introduced, for the first time, the organizational structure of conversational turn-takings. There are two components in the structure, the turn-constructional component (which are unit-types with which a speaker may set out to construct a turn) and the turn-allocation components (which determine who should seize the next turn).

In spoken dialogue systems, turn-taking is detected by finding short pauses (usually between 0.5 to 1 second (Ferrer et al., 2002)) and they indicate the current speaker's intent to yield the turn. Problems with this simple rule-based approach are: premature system engagement (e.g. interruptions), or alternatively long mutual silence events (Ferrer et al., 2002). A more flexible pause-based technique was proposed by Bell et al. (Bell et al., 2001), where task-related features were used to decide whether a pause is a hesitation or an intended turn-yielding signal.

Various linguistic cues have been found to be highly related to turn-taking transitions. Schlangen (Schlangen, 2006) studied the usage of prosodic features (describing the shape of the intensity and the fundamental frequency curve) and syntactic features (n-gram based) to predict whether the speaker will continue speaking or the turn will shift to a different speaker. Some of the features were manually annotated and cannot be calculated in real-time. Other linguistic cues, such as pitch levels (Ward et al., 2010) and intonation (Gravano and Hirschberg, 2011) have been found to play a key role in turn-taking regulation. On a different perspective, the study of de Ruiter el al. (Ruiter et al., 2006) revealed that only syntax and semantics cues are necessary to find the end of the speaker's turn.

### C. Turn-taking in embodied agents

Embodied agents include both virtual avatars and robotic platforms which mimic face-to-face conversations. Incorporating conversational capabilities from spoken dialogue systems, embodied agents can additionally produce and respond to nonverbal communication cues, such as facial displays, hand gestures and body stance (Cassell, 2000). There are mainly two problems in turn-taking on embodied agents: how to comprehend human's multimodal turn-taking communication cues, and how to control their own turn-taking behaviors.

Regarding the first problem (i.e., comprehension), certain modalities have been found to correlate with turn-taking intentions, such as posture shifts (Padilha and Carletta, 2003), haptic affordances (Chan et al., 2008), head motions (Ishii et al., 2015), gaze shifts (Ishii et al., 2014a) and eye blinks (Oreström, 1983).

Regarding the second problem (i.e., control), different architectures have been proposed to control the agent's turn-taking behaviors. The Furhat system (Skantze et al., 2015) was proposed to produce filled pauses, facial gestures, breath and gaze to deal with processing delays during tur-taking interaction. The CADENCE architecture was developed to manages robot's turn-taking actions including speech, gaze, gesture and physical manipulations (Chao and Thomaz, 2016). Also, the DiscoRT architecture was introduced to support engagement maintenance with gesture, gaze and speech communications (Nooraei et al., 2014).

### D. Collaborative turn-taking in physical tasks

Turn-taking has also been studied in the context of human robot interaction, where a robotic assistant and a human worker take turns to collaborate in a task. Some challenges are related to the use of the physical space, e.g., how to negotiate shared working spaces and objects/tools with humans through turn-taking.

In a robot-assisted assembly task, Calisgan el al. (Calisgan et al., 2012) studied the types and usage frequencies of implicit, nonverbal cues used for regulating turn-taking between human and robot. It was found that hand gestural cues play a dominant role as turn-ending cue, and they often occur together with lower body cues such as stepping back. Similarly, the CHARM project developed robotic assistants which work alongside human workers in a manufacturing environment (Hart et al., n.d.). In this setting, touch, gaze and robot's hesitation movements are explored as factors used to regulate turn-takings. Gaze was used by a robot to interpret human intentions. In such context, the robot plays an assistive role by handing construction pieces over to the human worker in a flexible and adaptive collaboration setting (Sakita et al., 2004). Gaze was also used to predict human intentions for anticipatory motion planning on a robotic servant (Huang and Mutlu, 2016).

On social robots, the timing in multimodal turn-taking (i.e., speech, gaze and gesture) was investigated through a collaborative Towers of Hanoi challenge with Simon robot (Chao and Thomaz, 2012). The Sandtray humanoid robot was displayed in the Science Museum in Milan (Italy) to interact with children on collaborative game solving tasks, and it was found that children adapted their behavior according to the robot actions without being told so (Baxter et al., 2013). Similarly, turn-taking interactions were shown to be emergent in a drumming game with a humanoid robot (Kose-Bagci et al., 2008). In the area of autism therapy, stereotypical gaze patterns of children with autism spectrum disorder were identified whiling interacting with a humanoid robot (Mavadati et al., 2015).

The research described focused on turn-taking process modelling and robot turn-taking behavior control, without explicitly predicting the end of human's turn. This paper describes a cognitive model to predict human operator's end-of-turn. More specifically, this paper makes the following contributions: 1) presents TTSNet, a computational framework for predicting human's turn-taking intentions during a physical human robot collaborative task; 2) presents a formal definition of the collaborative task and related turn-events for turn-taking analysis; 3) describes a design of a multimodal human-robot interaction system between surgeons and robotic nurses in the OR; 4) evaluates the proposed TTSNet framework in a simulated surgery dataset.

## III. PROBLEM FORMULATION

This section presents the formulation to define the turn-taking problem for the following analysis. More specifically, it covers the formulation for human robot collaborative task, the associated turn-events, the human sensing scheme and the turn-taking prediction algorithm.

### A. Collaborative task and turn-events definition

Consider the case when a human agent $H$ is working with a robotic agent $R$ on a collaborative task $\mathcal{W}$. $\mathcal{W}$ includes a series of subtasks $w_k^a$, which are conducted alternatively between $H$ and $R$. The subscript $k$ indicates subtask indexes (i.e., $k = 1,2,...,\mathcal{K}$) and superscript $a$ indicates the agent who is responsible for this subtask (i.e., $a \in \{H, R\}$). For example, $w_1^H$ is the first subtask which is taken care of by $H$, and could represent the subtask of human inserting a screw into a drilled orifice. Similarly, $w_2^R$ is the second subtask which is conducted by $R$, and could represent the subtask of robot delivering an assembly part. Thus, the collaborative task is formally defined as: $\mathcal{W} \triangleq \{w_k^a \mid a \in \{H, R\}, k = 1,2,...,\mathcal{K}\}$. The subtask $w_k^a$ is further defined as a 4-element tuple: $w_k^a \triangleq (g_k, \vec{u}_k, z_k^b, z_k^f)$. $g_k \in \mathcal{G}$ is the action label and $\mathcal{G}$ is the set containing all the action labels, such as delivering parts or retrieving tools. $z_k^b$ and $z_k^f$ represent the beginning and finishing time of this subtask $k$. $\vec{u}_k \triangleq \{u_{kj} \mid j = 1,2,...,|\mathcal{U}|\} \in \mathbb{R}^{|\mathcal{U}| \times 1}$ is the probability distribution of task's objects (e.g., tools or assembly parts) used in this subtask. $u_{kj}$ is the probability that object $j$ ($j = 1,...,|\mathcal{U}|$) is going to be used in subtask $w_k^a$, $u_{kj} \in [0,1]$ and $\sum_{j=1}^{|\mathcal{U}|} u_{kj} = 1$. $\mathcal{U}$ is the set containing all the objects potentially involved in this task (e.g., a drill and screwdriver in a manufacturing setting, or scalpel, retractor and scissors in a surgical setting). $|\mathcal{U}|$ returns the number of elements in this set, which is the total number of objects available in this task. The subtask $w_k^a$ is treated as the "atomic" component in the definition, since a turn only happens during the transition of two subtasks.

While a human is performing subtask $w_k^H$ and time goes on from $z_k^b$ to $z_k^f$, agent $H$ gets closer in finishing this subtask and the intent to give out the turn becomes more apparent. We focus on asymmetric turn-taking where robotic assistants need to predict the surgeon's turn-taking intentions. Thus, only the transitions from $H$ to $R$ are considered (i.e., $w_k^H$ to $w_{k+1}^R$). Each transition from $w_k^H$ to $w_{k+1}^R$ defines a turn-event $E_k \in \mathcal{E}^{give}$, in which a human is showing an unambiguous intention to give out the turn (denoted as $\mathcal{E}^{give}$). On the other hand, for most part of subtask $w_k^H$, the human is focusing on the current operation and show no intention to yield the turn. This period implicitly defines a turn-event $E_k \in \mathcal{E}^{keep}$, in which the human intends to keep the turn (denoted as $\mathcal{E}^{keep}$). Each turn-event $E_k \in \{\mathcal{E}^{give}, \mathcal{E}^{keep}\}$ ($k = 1,...,\mathcal{K}$) spans a window of time $[t_k^s, t_k^e]$, where $t_k^s$ indicates the starting time and $t_k^e$ indicates the ending time. The collaborative task $\mathcal{W}$, subtask $w_k^a$ and turn-events are illustrated together in Figure 4.

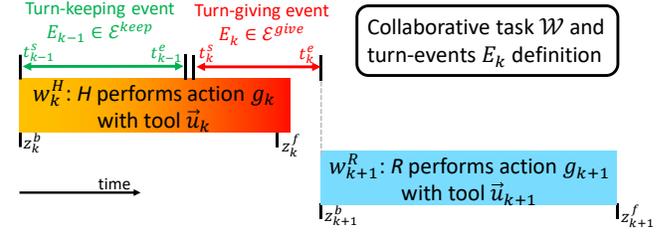

Figure 4. Illustration of the process when the human transits a turn to the robot. The definitions of collaborative task $\mathcal{W} = \{w_k^a\}$, the subtask $w_k^a = (g_k, \vec{u}_k, z_k^b, z_k^f)$ and the turn-events $E_k \in \{\mathcal{E}^{give}, \mathcal{E}^{keep}\}$ are demonstrated.

Under such definition, the moment when robot starts taking over the turn (i.e., $z_{k+1}^b$) is determined by the robot's estimate about the human's turn-giving intention, i.e., recognizing that human has already moved into state $\mathcal{E}^{give}$ from state $\mathcal{E}^{keep}$. Given an unknown turn-event $E_k$, sensor and context information within this event can be used to classify whether $E_k$ belongs to $\mathcal{E}^{give}$ or $\mathcal{E}^{keep}$. Such binary *End-of-Turn (EoT)* detection approach has been commonly adopted in turn-taking analysis (Arsikere et al., 2015; Bonastre et al., 2000; De Kok and Heylen, 2009; Guntakandla and Nielsen, 2015; Heeman and Lunsford, 2015; Saito et al., 2015; Schlangen, 2006).

### B. Human sensing

While the human operator is working on the subtask $w_k^H$, he/she is monitored through a collection of sensor readings $\vec{s}(t)$. The $M$ sensor channels $\vec{s}(t) = [s_1(t),...,s_M(t)] \in \mathbb{R}^{1 \times M}$ include physiological, physical and neurological signals, and are measured via various sensors (e.g., Kinect optical sensor, EEG sensor and EMG sensor). For example, $s_1(t)$ could represent the roll orientation of the human head.

Given a turn-event $E_k$ which spans time $[t_k^s, t_k^e]$, the sensed human signals $\vec{s}(t)$ within this time window are stacked to form a matrix representation of the human state:

$$X_k \triangleq [\vec{s}(t_k^s:t_k^e)] \in \mathbb{R}^{L_k \times M} \quad (1)$$

where $L_k$ is the length of event (i.e., $L_k = t_k^e - t_k^s$). For each stacked segment $X_k$, a label $y_k \in \{0,1\}$ is assigned to it to indicate whether the human wants to give out his turn (i.e., $y_k = 1$ when $E_k \in \mathcal{E}^{give}$) or keep the turn (i.e., $y_k = 0$ when $E_k \in \mathcal{E}^{keep}$) within this turn-event. An illustration of the multimodal sensing process and the matrix representation is shown in Figure 5.

### C. Predicting human turn-taking intention

The turn-taking intention estimation algorithm, referred as $\phi(\cdot)$, calculates an estimate of the turn-event type for $E_k$, given its multimodal sensing input $X_k$ i.e., $\hat{y}_k \triangleq \phi(X_k) \in \{0,1\}$. Moreover, the type of the turn-event $E_k$ should be recognized before the full event is completed (i.e., given only partial observations of $X_k$). This way, the human's intent to relinquish the turn can be recognized in an early stage and the robot can start moving early to facilitate this transition.

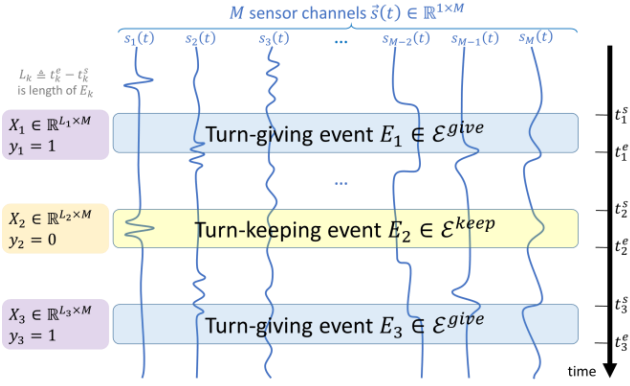

Figure 5. Illustration of the multimodal sensing process and the matrix representation for turn-taking analysis.

The parameter $\tau$ ($0 < \tau \leq 1$) is used to characterize the amount of partial observations to recognize a turn-event type. Given partial observations $X_k^\tau \in \mathbb{R}^{(\tau L_k) \times M}$ as the beginning $\tau$ fraction of full $X_k$, an early decision is made according to $\hat{y}_k^\tau = \phi(X_k^\tau) \in \{0,1\}$. An illustration of the early prediction scheme and the parameter $\tau$ is presented in Figure 6. The smaller $\tau$ is, the earlier this turn-giving intent can be recognized, but at the same time less accurate the algorithm becomes. The resultant dataset $\mathcal{D}^\tau$ is then used to evaluate the performance of the turn-taking intention estimation algorithm $\phi(\cdot)$, as the following:

$$\mathcal{D}^\tau = \begin{cases} X_k^\tau, X_k^\tau \in \mathbb{R}^{(\tau L_k) \times M} \\ y_k, y_k \in \{0,1\} \\ \hat{y}_k^\tau, \hat{y}_k^\tau = \phi(X_k^\tau) \in \{0,1\} \end{cases} \quad (2)$$

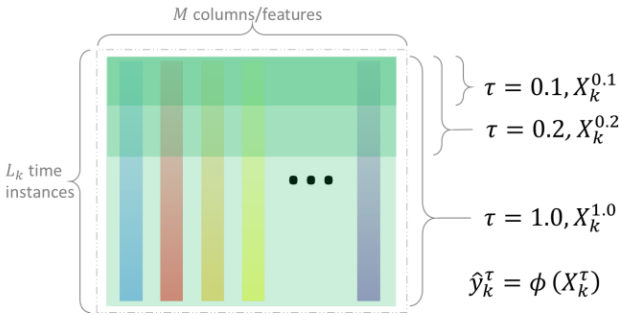

Figure 6. Illustration of the early prediction scheme and parameter $\tau$.

### D. Predicting turn-taking objects

The previous sections discussed the turn-taking intention prediction algorithm, which can recognize human's turn-taking intention before it is fully expressed. This would allow robots to start moving early to facilitate the upcoming turn-transition. However, at such an early stage, the robot often has not enough information about which objects (e.g., assembly parts or hand tools) are needed in the coming turn. Therefore, the "what" of turn-taking should also be addressed, i.e., predicting the most likely turn-taking object that is going to be used next. With this knowledge, the robot would be able to prepare the right object for the coming turn.

As defined in section III.A, the collaborative task $\mathcal{W}$ consists of $\mathcal{K}$ alternated human and robot subtasks $\mathcal{W} = \{w_1^H, w_2^R, w_3^H, w_4^R, \ldots, w_\mathcal{K}^R\}$. Each subtask $w_k^a$ consists of the probability distribution of objects used in this subtask, denoted as $\vec{u}_k = \{u_{kj} \mid j = 1, 2, \ldots, |\mathcal{U}|\} \in \mathbb{R}^{|\mathcal{U}| \times 1}$ where $u_{kj} \in [0,1]$ is the probability that object $j$ ($j = 1, \ldots, |\mathcal{U}|$) is used in subtask $w_k^a$, and $\mathcal{U}$ is the set containing all the objects. The function $F(\cdot)$ maps $w_k^a$ to its element $\vec{u}_k$, i.e., $\vec{u}_k = F(w_k^a)$. Therefore, given a collaborative task $\mathcal{W}$, its object sequence profile $U_k$ was constructed by stacking $\vec{u}_k$ column-wise, i.e.,

$$U_k \triangleq F(\{w_1^H, w_2^R, \ldots, w_k^R\}) = [\vec{u}_1, \vec{u}_2, \ldots, \vec{u}_k] \in \mathbb{R}^{|\mathcal{U}| \times k} \quad (3)$$

A sample $U_k$ is illustrated in Figure 7, given ten subtasks (i.e., $k = 10$) and an object set of five surgical instruments. The turn-taking object prediction algorithm, denoted as $O(\cdot)$, is used to predict the probability ($\vec{u}_{k+1}$) of each task object to be used in the following subtask, based on an observation of past object sequences ($U_k$), i.e., $\vec{u}_{k+1} = O(U_k)$. With this predictive result, the robot can then anticipate the most likely object to be used and preparatory movements can be executed in advance to facilitate the turn.

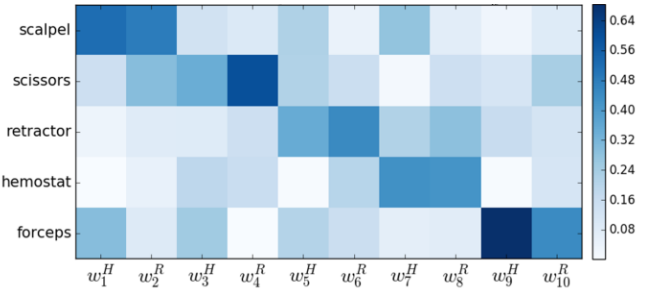

Figure 7. Illustration of a sample object sequence profile $U_k$. Cell intencity $u_{kj}$ represents the probability that object $j$ (on y-axis) is used in subtask $w_k^a$.

## IV. BACKGROUND IN SPIKING NEURAL NETWORKS

This section gives a brief overview of the background knowledge for spiking neural networks, in order to better understand the proposed TTSNet framework. A general introduction to SNN model is given in section IV.A, followed by a description of the SNN neural models and network structures in section IV.B.

### A. Spiking Neural Network Introduction

Conventional neural network models enforce synchronous firing of neurons of the same layer, as depicted in Figure 8. The connections between consecutive layers are forced to have the same conduction delay, thus all the neurons of the same layer can fire at the same time. This rigid structure poses difficulties when modeling multimodal temporal sequences, since the delays between layers are fixed and cannot adapt to different temporal resolutions associated with multimodal signals (Marcus and Westervelt, 1989).

SNN, however, can model the variability of axonal conduction delays between neurons. Because of the uniform conduction delays, the times for the firings to traverse the network will be different. This way, the asynchronous effect of multimodal signals can be modeled.

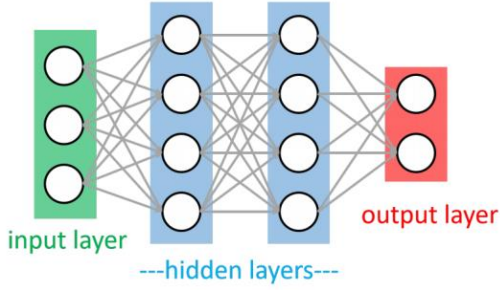

Figure 8. Conventional neural networks (i.e., Convolutional Neural Networks) have fix layers (i.e., input, hidden and output layers). Connections between layers have the same conduction delay and all the neurons of the same layer can only fire at the same time.

An illustration of a SNN is given in Figure 9. In Figure 9 (left) an example of a minimum spiking neural network with variable conduction delays is given. The numbers on the arrows indicate the required traverse time to arrive at the destination neuron. In Figure 9 (left), neurons $b,c,d$ fire at the same time (0 ms). Their responses arrive at neuron $a$ and $e$ at different times, resulting in insufficient potential to elicit the neuron. In Figure 9 (right), neuron $b,c,d$ fires at $\{2,0,1\}$ ms, respectively. They arrive at neuron $a$ at the same time, resulting in enough potential to elicit a potent post-synaptic response. Such behavior results in a time-locked pattern among neurons $\{c,d,b,a\}$, forming a PNG group which responds to this type of spatial-temporal pattern (i.e. neuron $b,c,d$ fires at $\{2,0,1\}$ ms, respectively).

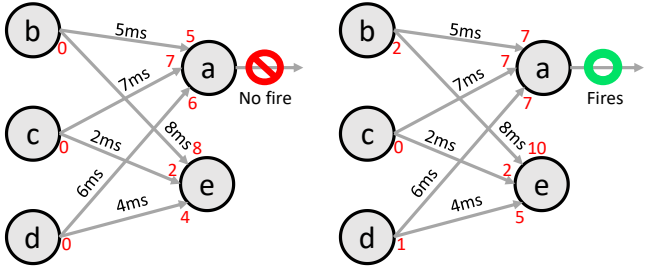

Figure 9. (left) Illustration of a minimal SNN, different synaptic connections have different conduction delays as indicated by the black numbers on the arrows. Red numbers indicate the fired time, the number at the end of the arrows indicates the spike arrival time at the post-synaptic neuron. Insufficient potential to elicit neuron $a$ (left). Neurons $b,c,d$ fire at a time locked pattern, generating enough potential to elicit neuron $a$ (right).

Turn-taking prediction using SNN requires a training process with two stages. The first stage trains the SNN network weights by feeding training data repeatedly into the network. Each training observation (i.e., feature vector) activates corresponding neurons in sequence, and the network weights are updated accordingly following a plasticity rule. The second stage of training consists of constructing salient patterns from training inputs for different classes. Those patterns form the signatures/templates for each class and are used for classification purposes. The testing phase includes feeding the unknown sequence into the trained SNN and getting the corresponding patterns, then comparing the similarity between the unknown data's pattern and the signature patterns of different classes. More details will be given in the following for each step.

### B. Neural Model and Network Structure

The underlying computational model for spiking neurons are introduced in this section. Also, the network structure which connects multiple spiking neurons together into a SNN is presented.

The basic model for the spiking neural model was originally introduced by Izhikevich (Izhikevich, 2006). The network has 250 neurons ($N = 250$), with 200 excitatory neurons (i.e., can be stimulated, $N_e = 200$) and 50 inhibitory neurons (i.e., cannot be stimulated, $N_i = 50$). Each excitatory neuron has 25 post synapses, connecting it to 25 other neurons (i.e., 10% of all neurons), following a uniform distribution. Each inhibitory neuron also has 25 post synapses, connecting it to 25 excitatory neurons following a uniform distribution. Each synapse has a conduction delay in the range of [1, 20] ms, following a uniform distribution. The conduction delay is the required amount of time for a signal to traverse through the synaptic connection. The weights of the synaptic connections are initialized to be +6 for all post synapses after excitatory neurons, and −5 for all post synapses after inhibitory neurons. Those weights represent how strong the synaptic connection is between two neurons, and are updated based on the Spike Timing-Dependent Plasticity (STDP) rule during the first stage of training phase. The maximum weight for each synaptic connection is set to 10.

The computational model which governs the firing/spiking behavior for each neuron is depicted by a two-dimensional system of ordinary differential equations (Izhikevich, 2003), as given in (4) and (5):

$$v' = 0.04v^2 + 5v + 140 - u + I \\ u' = a(bv - u) \quad (4)$$

where $v'$ and $u'$ represents first-order time derivative. The auxiliary after-spike resetting follows:

$$\text{if } v \geq +30 \text{ mV, then } \begin{cases} v \leftarrow c \\ u \leftarrow u + d \end{cases} \quad (5)$$

Here the variable $v$ represents membrane potential of the neuron and $u$ represents a membrane recovery variable which provides negative feedback to $v$. The variable $I$ is the input DC current to this neuron, which is set to +20mA when this neuron is stimulated based on input multimodal data. As illustrated by Figure 10, $a$ represents the time scale of the recovery variable $u$, $b$ represents sensitivity of the recovery variable $u$ to the subthreshold fluctuations of the membrane potential $v$. Also, $c$ represents the after-spike reset value of the membrane potential $v$, and $d$ represents after-spike reset increment of the recovery variable $u$. Depending on the four parameters $(a, b, c, d)$, this spiking neural model is able to reproduce spiking and busting behavior of known types of cortical neurons (Izhikevich, 2003). There are several different types of neuron kernels which can be used as the building block for excitatory neurons and inhibitory neurons (Connors and Gutnick, 1990; Gibson et al., 1999; Gray and McCormick, 1996),. Regular spiking (RS) firing patterns and fast spiking (FS) firing patterns have been commonly used for excitatory and inhibitory neurons (Izhikevich, 2003). However, there are other options which might suit the context of this problem better, such as Intrinsically Bursting (IB), Chattering (CH) and Low-threshold Spiking (LTS). The commonly observed neuron dynamics/types (Connors and Gutnick, 1990; Gibson et al., 1999; Gray and McCormick, 1996) and their corresponding parameters are shown in Table 1. Each neuron type belongs to either excitatory cortical cells (EX) or inhibitory cortical cells (IN), depending on their spiking pattern. The stereotypical firing patterns for these five neurons types are shown in Figure 10.

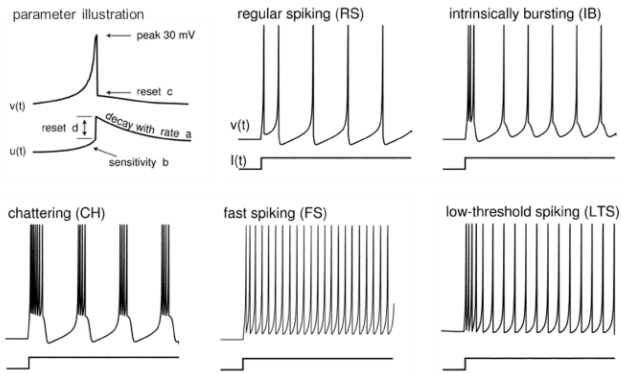

Figure 10. Known types of common neuron types and their simulation results based on the neural model. RS, IB and CH are excitatory neurons, FS and LTS are inhibitory neurons. Each subfigure shows voltage response of different neurons to a step of DC-current $I = 10$ mA. Time resolution is 0.1ms. Electronic version of the figure and reproduction permissions are freely available at www.izhikevich.com.

**Table 1. Different neuron dynamics and corresponding parameters**

| Neuron type | type | $a$ | $b$ | $c$ | $d$ |
|---|---|---|---|---|---|
| Regular Spiking (RS) | EX | 0.02 | 0.2 | -65 | 8 |
| Intrinsically Bursting (IB) | EX | 0.02 | 0.2 | -55 | 4 |
| Chattering (CH) | EX | 0.02 | 0.2 | -50 | 2 |
| Fast Spiking (FS) | IN | 0.1 | 0.2 | -65 | 2 |
| Low-threshold Spiking (LTS) | IN | 0.02 | 0.25 | -65 | 2 |

V. TURN-TAKING SPIKING NEURAL NETWORKS (TTSNET)

This section presents the TTSNet framework. Detailed descriptions are given for different aspects of the TTSNet framework, including neuron mapping (section V.A), SNN training (section V.B) and descriptive feature construction (section V.C).

*A. Neuron Mapping*

SNN can be used to predict turn-taking behaviors. For this, input multimodal data needs to be mapped to the neurons in the network. In previous research discrete input data was mapped to neurons on a one-to-one basis. For example, for hand-written digits' recognition, each pixel in the image (16×16) was mapped to one neuron in the network, resulting in a networks with 256 neurons (Rekabdar et al., 2016). The orientation of the written digits was mapped as nine orientations, which was assigned to fire five randomly chosen neurons in the network (Rekabdar et al., 2015b). However, mapping multimodal continuous-valued signals into SNN neurons requires a different approach because the size of the network grows exponentially as more features are introduced, and therefore the computation becomes intractable. Assume that the multimodal signal have $M$ channels, and each channel is quantized to have $V$ discrete levels and each level corresponds to five random neurons in the network (Rekabdar et al., 2015b). Then the resultant SNN will have $(5V)^M$ neurons to encode all the possible combination of inputs in the multimodal signal. In a small example of only five discrete levels (i.e., $V = 5$) and ten multimodal channels (i.e. $M = 5$), this would lead to $25^5 \approx 10^8$ neurons, which is intractable. This problem is solved by resorting to automatic channel quantization and decision-level fusion methods (Prabhakar and Jain, 2002).

First, each of the $M$ channels was quantized into $V$ levels. Data below 1-percentile and above 99-percentile is excluded to remove potential outliers. Then the $V$ bins are evenly distributed in the 1% ~ 99% range to encode the continuous sensor signals. Given 1% percentile value of $r_1$ and 99% percentile value of $r_{99}$, a given sensor reading value $s$ will be quantized to level $q$ ($0 \leq q \leq V - 1, q \in \mathbb{Z}$), following:

$$q = \begin{cases} 0, & s \leq r_1 \\ \dfrac{s - r_1}{r_{99} - r_1} V, & r_1 < s < r_{99} \\ V - 1, & s \geq r_{99} \end{cases} \quad (6)$$

The quantization process is applied to each of the $M$ channels of $X_k \in \mathbb{R}^{L_k \times M}$ and maps $X_k$ to $\tilde{X}_k \in \mathbb{Q}_V^{L_k \times M}$, where $\mathbb{Q}_V$ represents the quantized space with $V$ levels. Denote the partial observation as $\tilde{X}_k^\tau \in \mathbb{Q}_V^{(\tau L_k) \times M}$. For each quantized level of $q$, five excitatory neurons in SNN will be randomly allocated (mapped) following a uniform distribution. When level $q$ is active, all its five corresponding neurons will be stimulated one by one at 1 ms intervals, by providing a DC current of 20mA to variable $I$ in (1). The value of $V$ was set to be 40, to reach a total of $40 * 5 = 200$ excitatory neurons.

To deal with multimodal challenges, one SNN is constructed for each of the $M$ channels, and their final decisions are fused in the end. This approach follows the human brain mechanism for decision making (i.e., vision is not fused with hearing at a low-level, but is fused only after each modality is processed individually).

*B. SNN Training*

The training of the SNN includes two phases, the first phase adjusts the SNN synapses weights and the second phase identifies salient firing patterns for different classes of input. They will be detailed in the following.

*1) SNN Synapse Weight Training*

Once the mappings between input data $\tilde{X}_k^\tau$ and SNN neurons are established, the network needs to be trained. The training process consists of feeding relevant spatio-temporal patterns into the network and updating the synaptic weights based on Spike Timing-Dependent Plasticity (STDP) rules (Beyeler et al., 2013). Under STDP, the synaptic weights are updated based on the timings of the neural firings (Sjöström and Gerstner, 2010). The synaptic weights between those neurons which always fire together are strengthened. More specifically, the weight of synaptic connection from pre- to postsynaptic neuron is increased if the post-neuron fires after the presynaptic spike, i.e., the interspike interval $t > 0$. The magnitude of change decreases as $A^+ e^{-t/\tau^+}$. Reverse order results in a decrease of synaptic weight with magnitude $A^- e^{t/\tau^-}$. Parameters are set to $A^+ = 0.1, A^- = 0.12, \tau^+ = \tau^- = 20ms$, based on (Izhikevich, 2006). During this training stage, all the input patterns are mapped to their corresponding neurons in the SNN, and the synaptic weights are updated in each 1ms interval based on the STDP rules. Each quantized training data $\tilde{X}_k \in \mathbb{Q}_V^{L_k \times M}$ is fed into SNN for training and updating synaptic weights, following STDP rules. The time allocated to simulating each $\tilde{X}_k$ is 250ms ($T = 250$). Since the input data length $L_k < 40$ and each quantized level corresponds to five neurons, which are stimulated one at a time, the training pattern $\tilde{X}_k$ takes less than 200ms to stimulate the network. Then the network continues propagating the input without any active input, to allow the spike trains to propagate the network under STDP rules. Notice that patterns $\tilde{X}_k$ for both classes ($y_k \in \{0,1\}$) are presented to the SNN network during this training phase, following a random repeated order. The network is simulated for a total of 900s, which includes in total 3600 training inputs (some training inputs are fed into the model more than once), each of which takes 250ms to simulate. After the 250s simulation, the synaptic weights in the network do not change much. The difference between the synaptic weights of two consecutive frames has a 2-norm of 0.75, under a weight range of 10, so approximately 7.5% variation exists. Therefore, the simulation converges into a steady state.

*2) Signature Firing Maps Training*

The response of the trained SNN network given input $\tilde{X}_k$ is used for classification purposes. One SNN network is constructed for each information channel (i.e., one column of $\tilde{X}_k \in \mathbb{Q}_V^{L_k \times M}$, denoted as $\tilde{X}_{ki}$ for column $i$). Therefore, there will be in total $M$ SNN networks constructed, forming a SNN group. This is denoted as $\mathcal{S} = \{S_i\}, i = 1, ..., M$. Given input $\tilde{X}_k$, its group's response is denoted as $\mathcal{G}_k = \mathcal{S}(\tilde{X}_k)$. $\mathcal{G}_k$ consists of $M$ individual responses ($G_{ki}$) for each of the SNN networks, i.e., $\mathcal{G}_k \triangleq \{G_{ki}\}, i = 1, ..., M$ where response $G_{ki} \triangleq S_i(\tilde{X}_{ki})$. $G_{ki}$ denotes the Firing Maps (FM) when input $\tilde{X}_{ki}$ is presented to the model $S_i$, i.e., $G_{ki}$ encodes which neurons fired at what time. When shown an input $\tilde{X}_{ki}$ to the network, a simulation is created for $T$ milliseconds and each millisecond is the basic operation unit. There are in total $N$ neurons in the network which can be potentially fired. Therefore, $G_{ki}$ is formed as a $N$ by $T$ Boolean matrix (i.e., $G_{ki} \in \mathbb{B}^{N \times T}$), where a value of 1 at cell $(n, t)$ indicates that neuron $n$ ($1 \leq n \leq N$) fired at time $t$ ($1 \leq t \leq T$), and a value of 0 indicates no-firing, i.e.:

$$G_{ki}(n,t) = \begin{cases} 1 & \text{neuron n fired at time t} \\ 0 & \text{neuron n did not fire at time t} \end{cases} \quad (7)$$

The firing maps $\mathcal{G}_k = \{G_{ki}\}$ forms a compact and rich representation of the original signal $X_k$, and is used to predict the turn-taking type ($\hat{y}_k \in \{0,1\}$). When given partial observation $X_k^\tau$, its discretized version $\tilde{X}_k^\tau$ is fed into the SNN group $\mathcal{S}$, generating a partial response $\mathcal{G}_k^\tau = \mathcal{S}(\tilde{X}_k^\tau)$, which is used to predict its turn-taking type ($\hat{y}_k^\tau \in \{0,1\}$).

*C. Descriptive Feature Construction*

Features are constructed from $G_{ki}$ for turn-taking classification purposes. Because $G_{ki}$ is a large sparse matrix where most cells are zero, a more compact and effective feature representation is required. We propose the Normalized Histogram of Neuron Firings (NHNF) descriptors to compactly represent $G_{ki}$. More specifically, the total number of neurons (i.e. $N$) are evenly divided into $B$ bins, where bin $b$ ($b = 0, ..., B - 1$) covers neurons whose indexes are within the range $[bN/B, (b + 1)N/B]$. During a simulation of time $T$ ms, the number of total neuron firings corresponding to bin $b$ is counted and then divided by the simulation duration $T$ to generate the descriptor $h_k[i, b]$ for sample $X_k$ and feature $i$:

$$H_k \triangleq (h_k[i,b]) = \frac{1}{T} \sum_{t=1}^{T} \sum_{n=bN/B}^{(b+1)N/B} G_{ki}(n,t) \quad (8)$$

for $k = 1, ... K; i = 1, ..., M; b = 0, ..., B - 1$. Dividing the histogram by the simulation period $T$ makes this descriptor time-invariant, and thus is suitable for variable simulation lengths. This also allows the descriptor to be applicable to the early prediction problem, where the neuron firings of a partial time window is used instead of the entire duration $T$. Experiments with different bins ($B$) are shown in section VI.

$H_k$ consists of all the $M$ channels of information and the $B$ channels of histogram for a given sample $X_k$. $H_k$ is used to estimate the type of turn in the input ($\hat{y}_k$). When only partial responses $\mathcal{G}_k^\tau$ are available, the NHNF descriptors are extracted from it (denoted as $H_k^\tau$) and used to predict the turn-event type, i.e., $\hat{y}_k^\tau$.

*D. Turn-taking Object Prediction*

The turn-taking object prediction algorithm, denoted as $O(\cdot)$ above, gives the probability ($\vec{u}_{k+1}$) of each object expected to be used in the next subtask, based on an observation of past object sequences $U_k$, i.e., $\vec{u}_{k+1} = O(U_k)$. After the probability $\vec{u}_{k+1}$ is estimated, the most likely object to be used is given by *argmax* of all the probabilities, i.e.,

$$j_{k+1}^* = \underset{j=1,...,|\mathcal{U}|}{argmax} \; \vec{u}_{k+1}(j) \quad (9)$$

where $\vec{u}_{k+1}(j)$ represents the probability that object $j$ will be used in subtask $w_{k+1}^a$, and $j_{k+1}^* \in \{1, ..., |\mathcal{U}|\}$ indicates the most likely object to be used in subtask $w_{k+1}^a$.

A Hidden Markov Model (HMM) algorithm (Rabiner and Juang, 1986) was used to determine $O(\cdot)$. A HMM model $(\lambda_j)$ is characterized by three elements, the state transition probability $A_j$, the emission probability $B_j$ and the initial state $\pi_j$, i.e., $\lambda_j = (A_j, B_j, \pi_j)$. In total, $|\mathcal{U}|$ HMM models were constructed, one for each task objects, i.e., $\lambda_1, ..., \lambda_{|\mathcal{U}|}$. After the $|\mathcal{U}|$ HMM models were trained, they were used to estimate $\vec{u}_{k+1}$, the probability of objects required next. More specifically, $\vec{u}_{k+1}(j)$ is calculated by applying the *softmax* function on the fitting scores of the given observation sequences with each HMM model, i.e.:

$$\vec{u}_{k+1}(j) = \frac{e^{\mathcal{L}(\lambda_j; \widetilde{U}_k)}}{\sum_{l=1}^{|\mathcal{U}|} e^{\mathcal{L}(\lambda_l; \widetilde{U}_k)}}, j = 1, \ldots, |\mathcal{U}| \quad (10)$$

The likelihood $\mathcal{L}(\lambda_j; \widetilde{U}_k)$ describes how well a trained HMM model $\lambda_j$ fits a given observation sequence $\widetilde{U}_k$, and is calculated through the Forward-Backward algorithm (Baum and Eagon, 1967). The observation sequence $\widetilde{U}_k$ is generated by stacking the indices of the previously requested objects, one after the other (i.e., performing an *argmax* operation column-wise on $U_k$). With $\vec{u}_{k+1}$, the most likely object ($j^*_{k+1}$) can be estimated, and the robot is then able to prepare for the incoming turn-taking transition.

## VI. EXPERIMENTS

The turn-taking prediction algorithm was tested on a robotic scrub nurse, where the surgeon's turn-taking intentions must be predicted ahead of time. This section discusses the relevant aspects in the experiment setup, including surgical task setup (section VI.A), human sensing and signal processing (section VI.B) and finally computational experiments (section VI.C).

### A. Surgical Task Setup

A simulation platform for surgical operations was used to capture turn-taking cues on surgeons. The platform consists of a patient simulator and a set of surgical instruments to conduct a mock abdominal incision and closure task (Martyak and Curtis, 1976). In this collaborative task $\mathcal{W}$, the surgeon and nurse collaborate by exchanging surgical instruments. The surgeon performs the surgical procedure, and then the nurse searches, prepares and delivers the expected next surgical instrument.

Participants were recruited to perform the mock surgical task. Twelve participants were recruited from a large academic institution, with the age range of 20 to 31 years ($M = 25.7$, $SD = 2.93$). After inform consent was given (IRB protocol #1305013664), participants completed a training session. The participants were shown the surgical instruments ahead of time with their respective names, and a training video of step-by-step instructions of the mock abdominal incision and closure task (ten minutes) was shown to them. After seen the video, the participants had a "warm-up" trial. Each participant repeated the surgical task five times to reach a standard performance level. The training sessions and repeated trials led to a dataset of high face-fidelity which is realistic enough to validate the early turn-taking prediction capability as the core of this paper.

Each trial of the surgical task included in average 14 instrument requests. The surgical request actions were annotated as turn-giving events ($\mathcal{E}^{give}$), and the surgical operation actions were annotated as turn-keeping events ($\mathcal{E}^{keep}$). Participants in the role of "annotators" observed previously recorded videos of the surgical task. For each video presented, annotations were required consisting of the starting $t_k^s$ and ending time $t_k^e$ for each turn-event $E_k$, as well as the type of the segmented turn-event ($\mathcal{E}^{give}$ or $\mathcal{E}^{keep}$). The annotations were conducted by two human annotators (one main annotator annotated all, and a second annotator annotated 10% randomly selected segments) with a inter-rater reliability of Cohen's $\kappa = 0.95$ (Cohen, 1960). Overall, 846 turn-giving events ($y_k = 1$) and 1305 turn-keeping events ($y_k = 0$) are generated for turn-taking analysis.

### B. Multimodal Human Sensing and Processing

Communication cues were collected from the participants acting as surgeons during the simulated operation. Three sensors were used together to capture the multimodal signals, namely Myo armband, Epoc headset and Kinect sensor. Each sensor captures multiple channels of information, as illustrated in Figure 11. There are in total $M$ ($M = 50$) channels of information, and real-time signals from all $M$ channels were synchronized at a frequency of 20 Hz and concatenated column-wise (i.e., data at time $t$ corresponds to one row in $X_k \in \mathbb{R}^{L_k \times M}$).

Preprocessing techniques were used to smooth and normalize the multimodal signals. Each of the $M$ channels of information was first smoothed with Exponentially Weighted Moving Average (EWMA) approach, which is a common noise reduction technique for time-series data (Lucas and Saccucci, 1990). The weight for raw sensor measurement in EWMA was empirically set to 0.2. Then, each of the $M$ channels was normalized to have zero mean and unit variance, based on the grand mean and pooled variance from both turn events. This would then enforce the multimodal signals to be in similar magnitudes for further comparison and combinations.

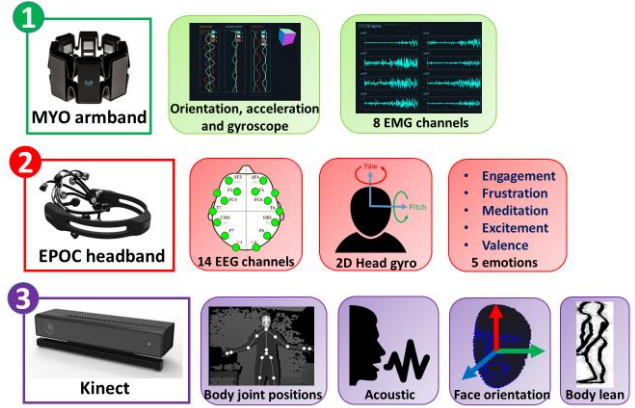

Figure 11. Multimodal human sensing. The three sensors are used simultaneously to capture human's multimodal signals.

The automatic feature construction and selection algorithm as proposed in (Zhou and Wachs, 2016) was used here. Each channel of the signal was first convolved with a filter bank containing six filters, i.e., identity transformation, Sobel operator, Canny edge detector, Laplacian of Gaussian detector and two Gabor filters, creating an encoded version of the signal. Then the correlation between each encoded signal with the turn-event labels was calculated through $\chi^2$ test of independence. Then the $m$ features of the largest test statistics values were retained as the final feature set, since a large value indicates high correlation with labels. In this experiment, the value of $m$ was empirically set to 10. This representation was quantized and mapped to SNN neuron based on the process described in section V.A. The top ten selected features are shown in Table 2.

**Table 2. Selected Top Features**

| Rank | Feature name + Filter name | $\chi^2$ |
|---|---|---|
| 1 | Epoc.gyro_y + identity | 1456.6 |
| 2 | Epoc.gyro_y + gabor1 | 1479.2 |
| 3 | Epoc.gyro_y + gabor2 | 1430.9 |
| 4 | kinect.audioConfidence + gabor1 | 1424.7 |
| 5 | kinect.audioConfidence + identity | 1408.5 |
| 6 | kinect.audioConfidence + gabor2 | 1388.0 |
| 7 | myo.orientation_x + gabor1 | 990.3 |
| 8 | myo.orientation_x + gabor2 | 975.9 |
| 9 | myo.acceleration_y + gabor1 | 975.1 |
| 10 | myo.acceleration_y + gabor2 | 971.1 |

*C. TTSNet Performance*

To evaluate the performance of TTSNet in predicting surgeon's turn-intentions, the following experiments were conducted. The experiment setup followed a leave-one-subject-out (loso) cross validation, where in each fold, the data from eleven subjects is used for training and the last subject's data is used for testing. Such evaluation scheme can evaluate the algorithm's generalization capability on unseen/novel subjects, and has been commonly adopted by the literature (Esterman et al., 2010). There are in total twelve participants in this study, therefore, there are in total twelve cross-validation folds. For accuracy measurement between prediction result $\hat{y}_k \in \{0,1\}$ and ground truth $y_k \in \{0,1\}$, the $F_1$ score (i.e., harmonic mean of precision and recall) was calculated.

The TTSNet can recognize the type of the turn-event, given only partial observation $X_k^\tau \in \mathbb{R}^{(\tau L_k) \times M}$. An early decision is made according to $\hat{y}_k^\tau = \phi(X_k^\tau) \in \{0,1\}$, based only on the beginning $\tau$ fraction ($0 < \tau \leq 1$). To evaluate the algorithm's performance in early prediction, the $F_1(\tau)$ for $\tau \in \mathcal{T}$ is calculated ($\mathcal{T} = \{0.1, 0.2, ..., 1.0\}$). Besides the point-wise $F_1$ scores, the Area Under the Curve (AUC) was also calculated to summarize the overall performance of a given curve. The AUC is calculated following:

$$AUC = \sum_{\tau \in \mathcal{T}} \Delta \tau * F_1(\tau) \qquad (11)$$

where $\Delta \tau$ is the step length equal to 0.1.

*1) Effect of Different Base Classifiers for TTSNet*

The purpose of this experiment is to compare the performances of different base classifiers when predicting the turn-event type $\hat{y}_k^\tau$ for a given unknown input $X_k^\tau$. The unknown partial input $X_k^\tau$ was first discretized into $\tilde{X}_k^\tau$ following the process described in section V.A. Then $\tilde{X}_k^\tau$ was fed into the trained SNN groups $\mathcal{S}$, generating the firing map $\mathcal{G}_k^\tau$, as described in V.B.2). Afterwards, the firing map $\mathcal{G}_k^\tau$ was used to calculate the NHNF descriptor, $H_k^\tau$, as detailed in section V.C. Lastly, $H_k^\tau$ was fed into a classifier to calculate $\hat{y}_k^\tau$. To control the experiment procedure, the spiking neural kernels were fixed so that we can focus on the effect of base classifiers only. The RS configuration is used for excitatory neurons and FS configuration is used for inhibitory neurons.

The final feature descriptor $H_k^\tau$ was first normalized so that each channel has zero mean and unit standard deviation. Then different classifiers were tested to compare their performances, namely Naïve Bayes (NB), Support Vector Machines (SVM), Decision Trees (DT), Random Forest (RF), Extra Trees (ET) and Adaboost (AB). The implementation was based on scikit-learn library (Pedregosa et al., 2011). The hyper-parameters for each classifier were chosen based on a local five-fold grid search approach. The training data was separated into five splits randomly, and the classifier was trained on four splits and validated on the fifth split. This process is repeated for each of the five splits, and the hyper-parameters generating the highest average validation scores were selected for usage. The trained classifier was then tested on the held-out test split. Following this process, the $F_1$ curves and the calculated AUC values are shown together in Figure 12. The SVM classifier generated the highest $F_1(\tau)$ scores for all $\tau$ points, and therefore leading to the highest AUC values. Therefore, it was the optimal classifier for our scenario and was used in the remaining analysis.

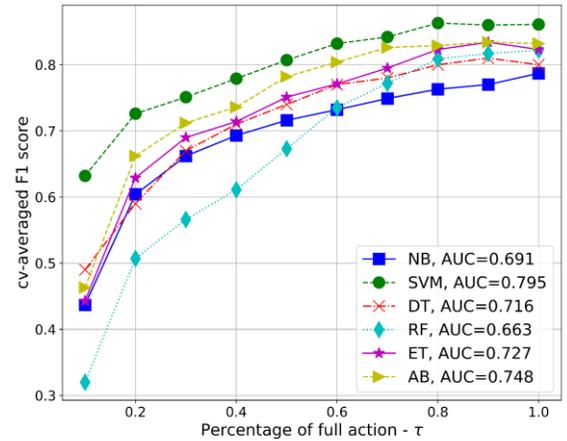

Figure 12. Effect of different base classifiers for TTSNet performances

*2) Effect of Different Neuron Kernels for TTSNet*

The purpose of this experiment was to compare the performances of different base neuron types. The detailed description of parameter configurations for each neuron type is given in Table 1. There are in total three types of excitatory neurons (RS, IB, CH) and two types of inhibitory neurons (FS, LTS), resulting in a total of six combinations for excitatory-inhibitory neuron pairs. To control the experiment focus, the SVM classifier, as the optimal classifier found by the previous experiment, is used as the base classifier for all six neuron pairs.

We conducted the experiment with all six neuron kernel configurations and the performances are shown in Figure 13. As revealed, the RS-LTS pair shows the best performances due to the highest AUC scores. The RS-FS pair is also comparable to RS-LTS with a slightly lower AUC score. The other four neuron configurations are worse than these two with a noticeable performance margin. Therefore, the RS-LTS neuron configuration was used in the following studies, as the best performing neuron-pair configuration.

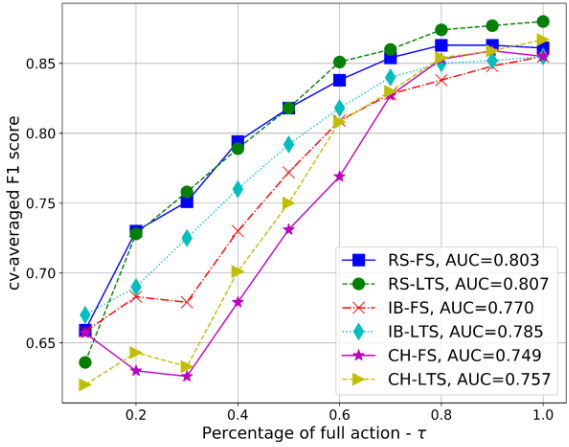

Figure 13. Effect of different neuron kernel types for TTSNet performance

*3) TTSNet vs the state-of-the-art*

The goal of this study was to compare the proposed TTSNet's performance against the state-of-the-art turn-taking prediction algorithms. To that end, different state-of-the-art algorithms have been implemented and tested on our dataset, as detailed below.

**Baseline 1: Hidden Markov Models (HMM)**

This baseline consists of a conventional temporal modeling algorithm, HMM, to predict turn-taking. HMM has been successfully used in several time-series modelling tasks such as turn-taking modelling (Zhang et al., 2006), gesture recognition (Jacob and Wachs, 2014) and speech recognition (Huang et al., 1990). In this scenario, one HMM model was trained for each of the two turn cases ($\lambda_0$ for turn-keeping and $\lambda_1$ for turn-giving). The training was based on the Baum-Welch algorithm (Dempster et al., 1977) and the weights in transition and emission matrices were acquired. During testing, the unknown input was fed into both HMM models, and the fitting score was used for classification. The label of the HMM model which has a higher fitting score was used as the unknown input's label. In the early prediction scenario, only the beginning $\tau$ partial observation was used to calculate the fitting score (log-likelihood) for each trained HMM model, following:

$$\hat{y}_u^\tau = \underset{q \in \{0,1\}}{argmax}\ log(\mathcal{L}(\lambda_q; X_u^\tau)) \qquad (12)$$

where $\mathcal{L}(\lambda_q; X_u^\tau)$ represents the log-likelihood of observation $X_u^\tau$ for HMM model $\lambda_q$, and was calculated through forward-backward procedure (Baum and Eagon, 1967). The hyper-parameters for HMM were selected empirically: five states, fully connected transitions and Gaussian emission models. The network was randomly initialized five times and the one generating the highest fitting score on a held-out validation split was selected in the end. The HMM implementation was based on the *hmmlearn* library (*hmmlearn*, 2017).

**Baseline 2: Ishii's approach**

This baseline (Ishii's) represents a state-of-the-art turn-taking prediction algorithm. Ryo Ishii proposed a set of turn-taking prediction frameworks to address the problem of turn-taking prediction in conversational settings (Ishii et al., 2016, 2015, 2014a, 2014b). Even though the application area is different from ours, his framework can still be adapted to work in this scenario as detailed in the following.

For normalization purpose, the original signal $X_k$ was normalized to the range of $[\mu_j - \sigma_j, \mu_j + \sigma_j]$ using a minmax scaler (Ishii et al., 2014b), for each of the $M$ channels of $X_k$. The mean $\mu_j$ and standard deviation $\sigma_j$ is the grand statistics calculated from all the training examples of channel $j$. After the normalization, each channel had an expected mean of 0.5 and a standard deviation of 0.5.

For feature extraction, the methods proposed in (Ishii et al., 2014b) and (Ishii et al., 2015) were used. The three features proposed in (Ishii et al., 2015) were used, namely average number of movement per second ($MO$), average number of amplitude per second ($AM$) and the frequency of movement per second ($FQ$). The eight descriptive statistics describing the shape of input signals as proposed in (Ishii et al., 2014b) are the min value ($MIN$), the max value ($MAX$), the amplitude of ranges ($AMP \triangleq MAX - MIN$), the duration of signal ($DUR$) and the slope of changes ($SLO \triangleq AMP/DUR$). A total of eleven features were constructed for each of the $M$ channels of $X_k$, and they were concatenated together to form the feature representation for $X_k$.

For classification purposes, the SVM classifier with RBF kernel was used, as suggested in (Ishii et al., 2014b). The hyper-parameters for the classifier were set based on a grid-search approach over logarithmic grids. More specifically, the error term penalty $C$ was searched over $\{10^0, 10^1, 10^2\}$ and the kernel coefficient $\gamma$ was searched over $\{10^{-1}, 10^{-2}, 10^{-3}\}$. The five-fold cross validation grid-search was conducted on the training spit only, and the hyper-parameters with the highest cv-averaged $F_1$ scores were selected as optimal.

For training, the features were extracted from the full observations $X_k \in \mathbb{R}^{L_k \times M}$ and then the SVM classifier was trained. For early prediction, given the partial observation of an unknown sample $X_k^\tau$, the same normalization and feature extraction processes were carried out on this partial observation, and the output of the SVM classifier forms the early estimate $\hat{y}_u^\tau$.

**Baseline 3: SNN-PNG**

This baseline is the SNN-based framework proposed in (Rekabdar et al., 2015a). The SNN-PNG framework was previously used for recognition tasks, such as hand-written digit recognition and gesture recognition. That framework is the most similar one to our proposed framework, with the major difference that they used polychronous neuronal groups (PNG) as features to encode the network output, and the k-nearest-neighbor (KNN) approach for classification. While in our framework, we directly extracted the NHNF descriptors from the neuron firing map. Also, instead of using the simple KNN classifier, the SVM classifier was used in our case. Additionally, our framework can deal with multimodal numerical inputs instead of discrete inputs.

The SNN-PNG framework is described in the following. As described in section V.B.2), the output of the SNN group to input $\tilde{X}_k$ is denoted as $\mathcal{G}_k$, which consists of $M$ individual responses ($G_{ki}$) for each channel. $G_{ki}$ consists of the firing map of input $\tilde{X}_k$ to network $S_i$. Within $G_{ki}$, the group of neurons fired together in a time-locked pattern forms a PNG $\mathcal{P}_k^i = \{P_{kt}^i\}$. When the map $G_{ki}$ fires, it shows the following sequence: neuron 4 and 7 fired together at 1ms (i.e., $P_{k1}^i = \{4,7\}$), then neuron 9 and 12 fired together at 7ms (i.e., $P_{k2}^i = \{9,12\}$), followed by neuron 11, 12, 47 at 9ms (i.e., $P_{k3}^i = \{11,12,47\}$), then the resultant PNG model for $\tilde{X}_k^i$ is $\mathcal{P}_k^i = \{P_{kt}^i\} = \{P_{k1}^i, P_{k2}^i, P_{k3}^i\} = \{\{4,7\},\{9,12\},\{11,12,47\}\}$. The collection of $M$ PNG groups $\mathcal{P}_k = \{\mathcal{P}_k^i\}, i = 1, ..., M$ encodes the spatial-temporal information embedded in $\tilde{X}_k$, and serves as the template for nearest-neighbor classification algorithms (Muja and Lowe, 2014). The PNGs from the turn-giving events (i.e., $\{\mathcal{P}_k|y_k = 1\}$) and turn-keeping events $\{\mathcal{P}_k|y_k = 0\}$ lead to a consensus of patterns that uniquely represent each class, and are used as the templates for each class.

For classification purposes, the K-Nearest-Neighbor (KNN) scheme was followed. An unknown input data $X_u$ was given and further discretized into $\tilde{X}_u \in \mathbb{Q}_V^{L_u \times M}$, then its PNG responses $\mathcal{P}_u$ were found. The classification task was then to find $\hat{y}_u \in \{0,1\}$ based on the similarity of $\mathcal{P}_u$ with $\{\mathcal{P}_k|y_k = 1\}$ and $\{\mathcal{P}_k|y_k = 0\}$ from the training examples. For distance measurement, the Jaccard index followed by Longest Common Subsequence (LCS) approach as proposed in (Rekabdar et al., 2016) was used. The Jaccard index measures the similarity between two PNG sets. For example, given two PNGs $P_{k1}^i$ and $P_{k2}^i$, the Jaccard index is defined as:

$$J(P_{k1}^i, P_{k2}^i) = \frac{|P_{k1}^i \cap P_{k2}^i|}{|P_{k1}^i \cup P_{k2}^i|} \quad (13)$$

and $J(P_{k1}^i, P_{k2}^i)$ is in the range of [0,1] where 0 indicates no similarity between set $P_{k1}^i$ and $P_{k2}^i$, and 1 indicates $P_{k1}^i = P_{k2}^i, P_{k1}^i \subseteq P_{k2}^i$ or $P_{k2}^i \subseteq P_{k1}^i$. $J(P_{k1}^i, P_{k2}^i)$ is then compared to a pre-defined threshold $J_\epsilon$ to be binarized into a value of 0 or 1. Then, the LCS algorithm was used to calculate the similarity $\sigma$ between two PNG groups $\mathcal{P}_m^i$ and $\mathcal{P}_n^i$, following:

$$\sigma(\mathcal{P}_m^i, \mathcal{P}_n^i) = \frac{LCS(\mathcal{P}_m^i, \mathcal{P}_n^i)}{\min(|\mathcal{P}_m^i|, |\mathcal{P}_n^i|)} \quad (14)$$

where $|\mathcal{P}_m^i|$ represents the length of the signature patterns $\mathcal{P}_m^i$. The $LCS(\mathcal{P}_m^i, \mathcal{P}_n^i)$ calculation is based on the binarized Jaccard index between $\mathcal{P}_m^i$ and $\mathcal{P}_n^i$. In order to classify an unknown input $X_u$, its PNG group response $\mathcal{P}_u$ was found and then compared with the training templates based on the $\sigma$ measurement. The average distance of $\mathcal{P}_u$ with $K_1$ turn-giving patterns was compared with the average distance with $K_0$ turn-keeping patterns. The closer cluster's label was then used as the label of the unknown pattern. The similarity measurement across the $M$ SNN channels were averaged to integrate the information together, following:

$$\hat{y}_u = \underset{q \in \{0,1\}}{\mathrm{argmin}} \frac{1}{M} \sum_{i=1}^{M} \frac{1}{K_q} \sum_{k=1}^{K_q} \sigma(\mathcal{P}_u^i, \mathcal{P}_k^i) \quad (15)$$

20 random training examples were selected from each class as the templates, i.e., $K_0 = K_1 = 20$. Three different values for Jaccard index threshold $J_\epsilon$ were iterated, namely $\{0.1, 0.5, 0.9\}$, and 0.9 was used due to its best performance. In the early prediction case, only the PNGs induced from the beginning $\tau$ fraction of input (i.e., $X_k^\tau$) was used for the classification.

**Baseline 4: Human**

This baseline reflects human performance when trying to predict upcoming turn transition points. A "button-press" paradigm was adopted here to measure human performance (Magyari et al., 2014). In this scheme, recorded videos of the surgical operation were played back to participants and then paused at random times. At every pause period, the participant was asked to guess what the surgeon's intent was (keep or relinquish the turn). The participants in this experiment used a cross-subject setting for data annotations (no self-annotation).

**Overall test**

The final plot for all the curves is presented in Figure 14. As shown, the proposed TTSNet outperforms all other state-of-the-art algorithms by a large margin, at every $\tau$ point. This result indicates the superiority of the proposed framework.

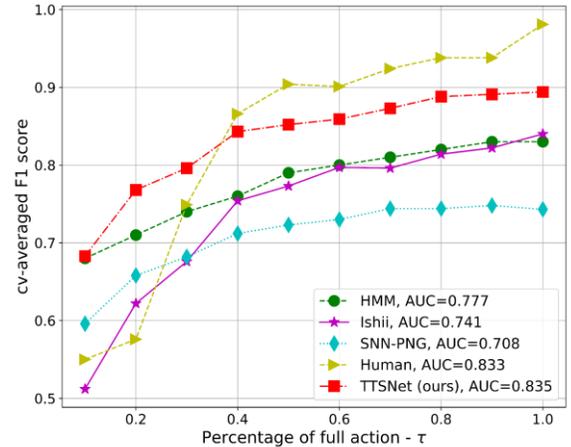

Figure 14. Comaprison with the state-of-the-art

*4) Relative important of individual feature*

The purpose of this experiment was to evaluate the performance when individual feature is used for turn-taking prediction. The selected top ten features as described in section VI.B were used. For each feature $i$ ($i = 1, ..., 10$), the corresponding SNN network $S_i$ and the extracted NHNF features were used. The SVM classifier and the RS-LTS neuron kernels were used. The cv-averaged $F_1$ score for each individual feature (and the ten features used together) is shown in Figure 15. As shown, a general trend is that a feature ranked higher (smaller number index) can generate better performances. The best performance was achieved when all ten features are used together.

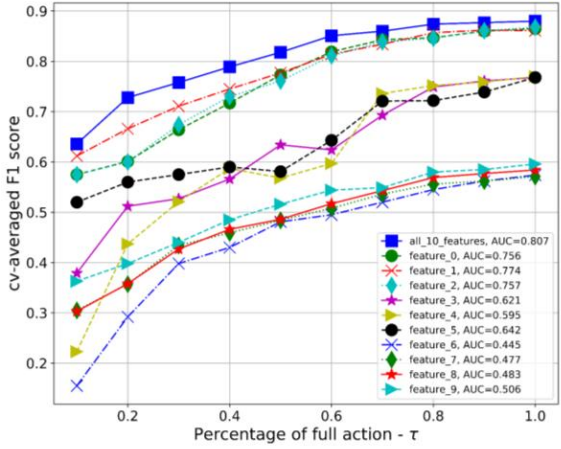

Figure 15. Performance of individual feature for turn-taking prediction

*5) Visualization of SNN Responses*

A visual representation of SNN responses to inputs of different classes is given here. Such visualization can give an intuition of what the neural network has learned to achieve effective turn-taking prediction. Figure 16 shows 6 neurons firing maps for each class of input. The SNN corresponding to the first feature was selected here for visualization. The responses to turn-keeping inputs ($\mathcal{E}^{keep}$) are on the top two rows, and the responses to turn-giving inputs ($\mathcal{E}^{give}$) are at the bottom two rows. Each response is represented by a neuron firing maps of the SNN, where a dot at location $(x, y)$ indicates that neuron $y$ fired at time $x$. The simulation lasts 250ms for each input, thus the x-axis ranges from 1 to 250. There are in total 250 neurons, thus the y-axis ranges from 1 to 250, indexing all the neurons (the first 200 neurons being excitatory and the last 50 neurons being inhibitory). As shown, the SNN responds differently to $\mathcal{E}^{keep}$ and $\mathcal{E}^{give}$ inputs.

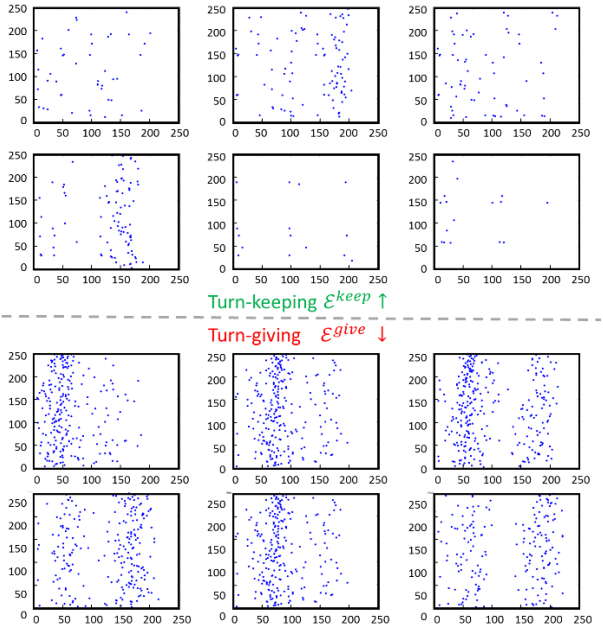

Figure 16. Comparison of SNN responses to different classes of input. The SNN firing map is shown for the turn-keeping inputs (top two rows) and turn-giving inputs (bottom two rows). X-axis is time (ms) and Y-axis is fired neuron index.

*6) Turn-taking Object Prediction Performance*

Here the proposed turn-taking object prediction algorithm is evaluated. The purpose of the task-object prediction algorithm $O(\cdot)$ was to predict the most likely instrument to be requested next ($j_{k+1}^*$), from an observation sequence $\widetilde{U}_k$. The design of $O(\cdot)$ is data-driven and relies on the HMM architecture. For that purpose, we used the instrument dataset collected through the same abdominal incision and closure task, as described in section VI.A. This task requires the usage of six different types of surgical instruments, namely scalpel, forceps, retractor, scissors, hemostat and needle. These six instruments were mapped to integers $\{1, \dots, 6\}$, respectively, in the object set $\mathcal{U}$. A concrete example is given to illustrate this. If the surgeon has just used scalpel, forceps and needle, and needs a hemostat in the next step, then the observation sequence $\widetilde{U}_k$ would be (1,2,6) to represent the past instrument IDs, and $j_{k+1}^* = 5$ to represent the hemostat to be used.

About 14 instruments were used in each trial of the abdominal incision and closure task. Even though this surgical task has clear goal and a relatively structured procedure, there were some sequence variations due to subjective preferences. Each subject performed the task five times, and in total twelve participants were recruited. That resulted in a total of 846 instrument requests, which served as the basic dataset for the turn-taking object prediction algorithm development.

We segmented the instrument sequences into smaller chunks for training. A trigram approach was used (Paliwal et al., 2014), i.e., $\widetilde{U}_k$ consists of the three surgical instruments requested prior to the current (e.g., $\widetilde{U}_k = (1,2,6)$ and $j_{k+1}^* = 5$). Different values of n-grams were tested (including 1,2,3,4,5) and 3 was found to be optimal. For those instrument requests happening in the beginning, zero-padding was used to construct $\widetilde{U}_k$, indicating an unknown previous instrument request. For example, to predict the third requested instrument when only two instruments were requested previous to it, the observation $\widetilde{U}_k$ would be (0,2,1) to indicate a missing value in the beginning.

The $|\mathcal{U}|$ HMM models ($\lambda_1, \dots, \lambda_{|\mathcal{U}|}$) were trained separately. For each HMM model $\lambda_j$, the observation sequences corresponding to this model were used to train its parameters $(A_j, B_j, \pi_j)$, using the Baum-Welch algorithm (or known as EM algorithm) (Dempster et al., 1977). Since the Baum-Welch algorithm can only find local maximum, the HMM model was randomly initialized 10 times, and the one generating the highest fitting score on a separated validation set was selected as the final model. This local validation set, which is different from the final testing split, was also generated following the loso principle (Esterman et al., 2010). The number of HMM states were selected based on a grid search over $\{1, 3, 5, 7, 9\}$, and it was found that the performances reached a plateau after five states. We utilized a fully-connected 5-state HMM structure, where a state can transit to any other state. This structure outperformed the left-right model in our case.

In total, six HMM models were trained, one for each instrument class. Since the six instruments have different usage frequencies in this surgical task, the number of training examples for each class is different. To compensate for the unbalanced class ratio in performance evaluation, random over-sampling technique was used (Liu et al., 2007) so that all the classes have the same number of training examples. The experiment follows the same loso cross validation setup, where a single subject's data was left out for testing. The performance of the proposed object prediction algorithm was compared with that of other classification algorithms as benchmarks, namely Naïve Bayes (NB), Support Vector Machines (SVM) with both linear kernels and rbf kernels, Decision Trees (DT) and Random Forests (RF) on the same observations $\widetilde{U}_k$. The hyper-parameters for each classifier were chosen based on a grid search over log-linear spaces. The classifier yielding the best cross-validation performances was chosen to be tested on the held-out test split.

The weighted $F_1$ score was used to evaluate the performance of the task object prediction algorithm. To calculate this metric, first the $F_1$ score was calculated individually for each class, then all the six $F_1$ scores were averaged together using the number of examples as weight. Due to the large variation of $F_1$ scores from all twelve cv-folds, we used the median of the twelve $F_1$ values to summarize the overall performance. The Median Absolute Deviation (MAD) is used as a robust measurement of variability in the $F_1$ scores, and is calculated as the median of the absolute deviations from the data's median: i.e.,

$$MAD = median(|\vec{F}_1 - median(\vec{F}_1)|) \quad (16)$$

The median and MAD metric for different benchmark algorithms and the proposed one are shown in Table 3.

**Table 3. Performance of object prediction**

| Algorithm | Performance (median ± MAD) |
|---|---|
| Naïve Bayes | 0.766 ± 0.017 |
| Linear SVM | 0.877 ± 0.010 |
| RBF SVM | 0.888 ± 0.037 |
| Decision Trees | 0.900 ± 0.015 |
| Random Forest | 0.899 ± 0.018 |
| Proposed (HMM) | **0.932** ± 0.010 |

As shown by the table, the proposed HMM-based turn-taking object prediction algorithm can achieve the best performance, compared to the benchmarks; a performance of 0.932 indicates that the proposed algorithm can predict the next turn-taking object with high accuracy.

## VII. DISCUSSIONS

### A. Discussion of base classifier selection

When evaluating the performance of different base classifiers for TTSNet, it was found that SVM yields the best performance, with an average $F_1$ score 5% higher than the second place (AB). SVM has been widely used as the underlying classifier in several other turn-taking recognition frameworks (Jokinen, 2010; Jokinen et al., 2013; Kawahara et al., 2012), and a similar finding is observed in this experiment.

### B. Discussion of neuron kernel selection

With respect to the different neuron kernels for TTSNet, it was found that RS-FS and RS-LTS pairs showed the best performances. The selection of exhibitory neuron kernel dominates the final performance. As long as the excitatory neuron type is RS, the selection of different inhibitory neuron types does not make a big difference. The IB neuron group (IB-FS and IB-LTS) achieved the second-best performance, while the CH neuron group (CH-FS and CH-LTS) showed the worst performance. In the CH neuron group, the performances even decrease as longer observations are given. This indicates that the selection of neuron kernel types is important in achieving the optimal performances when using the TTSNet framework. In our scenario, the RS kernel is found to be the most suited neural kernel to model the underlying spatio-temporal patterns for turn-taking. This is not a surprising, as RS is the most common excitatory neuron in mammalian neocortex (Izhikevich, 2004) and therefore is able to model a wide range of human behaviors, including turn-taking.

### C. Discussion of performance against state-of-the-art

When comparing the performance of TTSNet against the state-of-the-art turn-taking algorithms, it was found that the proposed TTSNet achieved the best performance out of all algorithms. HMM, as one of the strong baseline used as the core sequence modelling algorithm in other turn-taking frameworks (Zhang et al., 2006), achieved the second-best performance. The Ishii's framework, designed for conversational turn-takings, achieved the third-best performance. The worst-performed baseline is SNN-PNG approach, which is the most similar algorithm to ours (TTSNet). The only difference between SNN-PNG and TTSNet is that SNN-PNG uses PNG as features and relies on nearest-neighbor classification, while our approach relies on the proposed NHNF features and SVM for classification. This result indicates that the careful design and adaptation of SNN is important in achieving the best performance in turn-taking modelling, and simply using a previously proposed SNN framework cannot deliver optimal performances.

### D. Discussion of performance against human baseline

The proposed algorithm is found to yield better performance when compared against the human baseline, when little partial observation is given ( < 40% ). This behavior is partially due to the suitability of SNN for early prediction, since it can ignite the entire network from only a few anchor neurons in the beginning (Rekabdar et al., 2015b). When an anchor neuron fires in SNN, it generates a sequence of signals to traverse through a network, causes a spike train and continues to activate a group of neurons. This cognitive behavior enables the proposed SNN-based TTSNet framework to be capable of predicting human's turn-taking intentions at an early stage. Similar early prediction behavior of SNN have been noticed in hand digit recognition tasks (Rekabdar et al., 2016, 2015b) and gesture recognition (Botzheim et al., 2012).

### E. Discussion of feature importance

It was found that when the features were used individually in the TTSNet, performances can be grouped into three different groups of features. Feature 0,1,2 (referred as group A) performed similarly, feature 3,4,5 (referred as group B) performed similarly and the rest four features (referred as group C) performed similarly. Group A includes different encodings of the same source of information, thus, the performances were similar within the group. The same explanation follows for group B. Group C included features capturing forearm posture and gesture information (orientation and acceleration), and therefore the performances were similar. Notice that in the feature ranking procedure as described in section VI.B, the features from group A were ranked higher than those from group B, followed by group C. Here a similar trend is observed. The performance of group A is better than group B, which is then better than group C. Such observation provides evidence to support the feature selection methods as described in section VI.B. Also, notice that the best performance is achieved when all ten features were used together. This shows the power of using multimodal against unimodal interaction for turn-taking prediction.

*F. Discussion of SNN visualization*

When visualizing the learned SNN responses from TTSNet, it was found that different turn-taking events have different stereotypical SNN responses. The $\mathcal{E}^{give}$ inputs in general can fire more neurons in the trained SNN network compared to $\mathcal{E}^{keep}$ inputs, due to the larger firing intensities (reflected by the amount of points in $\mathcal{E}^{give}$ compared to $\mathcal{E}^{keep}$). This could mean that humans exhibit a coherent pattern when relinquishing their turn. The neurons in the TTSNet framework fire in the presence of such pattern. Another observation is that responses in $\mathcal{E}^{give}$ generally have a column-wise structure (either one column or two columns). This structure is generated when a group of neurons fire together in a time-locked pattern, forming a PNG as a signature of early turn-taking intent.

*G. Discussion of turn-taking object prediction*

The turn-taking object prediction experiment revealed that the proposed HMM-based algorithm can accurately predict the next turn-relevant object. A more detailed examination of the confusion matrix indicated that most errors came from a confusion between hemostat and needle. This is due to an intrinsic confusion in the surgical procedure. Towards the end of the abdominal incision and closure task, the surgeon would request multiple hemostats to open and stabilize the opening, followed by requesting a needle for suture. Depending on the situation of the tissue and the size of the opening, the surgeon would request two, three, four or even more hemostats. Therefore, after requesting three hemostats, the surgeon might request another hemostat or a needle. In order to solve this problem, other features need to be included to resolve the confusion between the two cases.

## VIII. CONCLUSIONS

In human robot interaction scenario, turn-taking capability is a critical component to enable robots to interact seamlessly, naturally and efficiently with humans. However, current turn-taking algorithms cannot help to accomplish early prediction. To bridge that gap, this paper proposes the Turn-Taking Spiking Neural Networks (TTSNet), which leverages cognitive models to achieve early turn-taking prediction. More specifically, this model is capable of reasoning about human's turn-taking intentions, based on the neurons firing patterns in a Spiking Neural Network (SNN). The TTSNet framework relies on multimodal human communication cues (both implicit and explicit) to predict whether a person wants to keep or relinquish the turn. Such decision can then be used to control robot actions.

The proposed TTSNet framework was tested in a surgical context, where a robotic scrub nurse predicted surgeon's turn-taking intentions in order to determine when to deliver surgical instruments. The algorithm's turn-taking prediction performance was evaluated based on a dataset, acquired through a simulated surgical procedure. The proposed TTSNet framework can achieve better performances than its counterparts. More specifically, the algorithm results in a $F_1$ score of 0.683 when 10% of complete action is presented, and a $F_1$ score of 0.852 when 50% of complete action is given. Such early prediction capability is partially due to the suitability of cognitive models (i.e., SNN) for early prediction. Such behavior would enable robots to perform turn-taking actions in an early stage, in order to facilitate the transition and increase the overall collaboration efficiency and smoothness.

There are some limitations of this work. The proposed TTSNet model was trained on a dataset collected in a simulated setting. When being used in a real OR, the model needs to be fine-tuned to adapt to the new setting. On the other hand, the turn-object prediction algorithm cannot generalize to cases when innovative surgical procedures are conducted and/or when unseen surgical instruments are used.

Future work includes 1) proposing a more comprehensive human state definition beyond only the two cases, 2) including more contextual information besides only multimodal signal (e.g., phase within a task and current task progress) to improve early prediction capability, 3) and also validating the proposed TTSNet framework in other scenarios beyond OR, such as robot-assisted manufacturing, robot companion and rehabilitation.